\documentclass[conference,10pt,letterpaper]{IEEEtran}

\usepackage{cite}

\usepackage{stfloats}

\ifCLASSINFOpdf
 \usepackage[pdftex]{graphicx}
\else
\fi
\usepackage[cmex10]{amsmath}
\usepackage{array}
\usepackage{mdwmath}
\usepackage{mdwtab}
\usepackage{eqparbox}
\usepackage{tikz}
\usepackage{amssymb}
\usepackage{amsthm}
\usepackage{subfig}
\usepackage{latexsym}
\usepackage{inputenc}
\usepackage{url}
\usepackage{euler}
\usepackage[inline]{asymptote}
\usepackage{proof}
\usepackage{tabularx}

\begin{document}

\title{A Study of FOSS'2013 Survey Data Using Clustering Techniques}
\author{\IEEEauthorblockN{A. Mani}
\IEEEauthorblockA{Department of Pure Mathematics\\
University of Calcutta\\
9/1B, Jatin Bagchi Road\\
Kolkata(Calcutta)-700029, India\\
Email: {a.mani.cms@gmail.com}\\
Homepage: \url{http://www.logicamani.in}}
\and
\IEEEauthorblockN{Rebeka Mukherjee}
\IEEEauthorblockA{Department of Comp.Sci.and Engg.\\
Netaji Subhash Engineering College\\
B-13 Rajdanga Nabapally\\
Kolkata-700107, India\\
Email: {rebekamukherjee@gmail.com}}}

%\titlerunning{Rough Counting}

\maketitle

\begin{abstract}
FOSS is an acronym for Free and Open Source Software. The FOSS 2013 survey primarily targets FOSS contributors and relevant anonymized dataset is publicly available under CC by SA license. In this study, the dataset is analyzed from a critical perspective using statistical and clustering techniques (especially multiple correspondence analysis) with a strong focus on women contributors towards discovering hidden trends and facts. Important inferences are drawn about development practices and other facets of the free software and OSS worlds.
\end{abstract}

\textbf{Keywords}: \begin{small}{Free Software, Survey, GNU/R, Women in Free Software, Clustering, FOSS 2013,  }\end{small}

\section{Introduction}

In the year 2002, the GSyC/LibreSoft research group at the Universidad Rey Juan Carlos conducted a survey of over 2500 contributors (and some users) to FOSS projects. The group has been involved for long in research and collaboration on Free and Libre Open Source Software projects. The FOSS'2013 survey \cite{floss2013} is similar to the earlier survey in that similar questions have been asked. The free software world has come a long way over the last fifteen years. The number of people involved in free software and OSS has grown tremendously and almost every IT company and sane Government invests heavily in free software related infrastructure. Contributors to free software projects are known to do so for diverse reasons ranging from commitment to freedom of knowledge to the purely commercial. 

Mainstream media has its commercial biases and is unlikely to report properly on the state of affairs in the free software world. Surveys of this kind are therefore essential for understanding the true state of affairs on the question of freedom. Since the last edition of the survey, free software and OSS communities have become more diverse from the point of view of project types, project size, skill set and background of contributors. It can be claimed that the FOSS'2013 survey failed to handle the entire demography in diversity and volume (for details see \cite{AIS2011} and related references). The purpose of this study is not to answer this, but to look at what can be inferred from the data set in question. 

For the basic definition of free software, OSS and related philosophy, the reader is referred to \cite{RMS}. The Geek feminism wikia can be referred to for a general overview of women's issues in the FOSS world. A shorter summary can be found at \cite{AM096}.

\subsection{Issues with the Dataset}

From a sampling survey perspective, the Internet based design of the survey has been far from perfect. The actual population of free software contributors is huge (the Linux kernel project itself had over 2200 developers in 2013, while CMS like Joomla have many times that many developers) and the number of respondents to the survey is less than 2200. The reasons for this is likely due to poor publicity, less familiarity with the research group, contributors being overwhelmed by too many free software related surveys, contributors becoming too focused on their own projects and experiencing too many deadlines. Unfortunately similar surveys with larger sample sizes are not known as of this writing. Though it is true that most if not all free software contributors may be contacted through the Internet, only a small dedicated fraction uses IRC these days. Publicity for the survey in non IRC channels was clearly limited.

Because of the survey design only descriptive statistical techniques and simple soft techniques could be used to study the data. Projections and models based on the data cannot be relied upon. Further constraints on the demography include the following:
\begin{itemize}
\item {$25\%$ of those surveyed work or reside in the USA, $8\%$ in Germany, $6\%$ in the U.K., $6\%$ in Spain, and $ 4\%$ in Australia.}
\item {About $40\%$ of those surveyed were native speakers of the English language, while over $72\%$ were at least proficient in the language.}
\item {Over $72\%$ of those surveyed had at least completed a bachelors degree at a university.}
\end{itemize}

\subsection{Description of Data Set}

Majority of the participants were developers and the anonymized (synthetic) dataset has 262 columns and 2183 rows. $53$ questions with many subquestions were put to participants. Except for a couple of columns all of the data is in factor form. Number of levels associated with columns are obviously question-dependent and do not admit of easy automatic conversion into numeric values.  

\section{Methodology}

The methodology used for analyzing the data was determined by the issues with the survey mentioned above and the nature of the data. The questions were split into the following groups:
\begin{itemize}
\item {First block of 42 questions of a personal nature}
\item {Questions $43-109$ relating to participant's general outlook in relation to free software and OSS worlds. This will be referred to as the \emph{Outlook} block.}
\item {Questions $110-168$ relating to participant's choices related to programming. This will be referred to as the \emph{Programming} block.}
\item {Questions $169-262$ relating to participant's experiences in and views about the free software and OSS development environment. This will be referred to as the \emph{Environment} block.}
\item {The above were further subsetted to analyze the involvement of women.}
\end{itemize}

The techniques that have been used in the analysis include descriptive statistical methods, multiple correspondence analysis (MCA) and clustering. All of the analysis has been done in GNU/R with related libraries.  MCA \cite{FSJ2011} can be viewed as an adaptation of principal component analysis to factor data in which categories are looked for by both columns and rows simultaneously. 

\section{Results}

First we mention some simple results derived from the data for providing some perspective:
\begin{itemize}
\item{$34\%$ consider themselves as a part of the Free Software community, while $32\%$ consider themselves part of the OSS community and the rest do not care about labels.} 
\item {$15\%$ of those surveyed think that Free Software and Open Source communities are not only different way of thinking but also different way of living. $29\%$ think that they are different only in the principles but they work in the same way while $21\%$ do not care.}
\item {About $51\%$ have been involved in $1-5$ FLOSS development projects, $18\%$ in $6-10$ projects, $7\%$ in $11-20$ projects, $2\%$ in $21-30$ projects, $1.6\%$ in $31-50$ projects, and $1.6\%$ in more than $50$ projects.}
\item {About $20\%$ of participants were involved in a project as a leader, coordinator or administrator in FLOSS development projects, $12\%$ in two projects, $7\%$ in three projects, $8\%$ in $4-5$ projects and $6\%$ in more than five projects in similar capacities. $29\%$ of the participants had never been involved as a leader, coordinator or administrator in FLOSS development projects.}
\item{Questions comparing the contribution of developers to the benefits they get from the FOSS community were ignored by nearly $50\%$ of the participants. Nearly $70\%$ of the participants believe that most developers are less concerned about money. }
\item{$2\%$ of the people surveyed totally agree that to develop FLOSS is a kind of self-exploitation, because people do not receive any benefits in return for good ideas and work,  $6\%$ agree, while $27\%$ disagree and $36\%$ totally disagree.} 
\item {Surprisingly only $51\%$ of participants used at least two programming languages.}
\item {About $29\%$ of Python Programmers know C as well.}

\end{itemize}

Some women specific information deducible from the data are as below:
\begin{itemize}
\item {$35\%$ of women participants were single, about $11\%$ do not live with their partners, while $3\%$ live with their partners. $3\%$ are married and $0.09\%$ are separated from their partners.}
\item {$20\%$ of women participants surveyed had children, while $79\%$ did not.}
\item {About $5\%$ of the participants were unemployed.}
\item {$40\%$ of the women participants work or reside in the USA, $6\%$ in Germany, $6\%$ in the India, $5\%$ in the U.K., and $4\%$ in Australia. To put this in perspective, the number of women respondents from India appears to be very low as the first author herself knew more than twenty women contributors in 2013.}
\item {$81\%$ of women contributors are at least graduates. Number of Ph.Ds form $10\%$ of the pool.}
\item {$82\%$ of women contributors are at least proficient in the English language.}
\end{itemize}

\begin{itemize}
\item {$32\%$ felt that they love their present job, while $5\%$ wanted a more interesting job than their present one.}
\item {$28\%$ consider themselves as a part of the Free Software community, while $4\%$ consider themselves part of the Open Source community.}
\item {$10\%$ of women participants think that Free Software communities are not only different way of thinking but also different way of living. $20\%$ think that they differ only in their principles but work in the same way. $21\%$ do not care.}
\end{itemize}

\subsection{Women Contributors}

$226$ of the $2183$ persons in the dataset identified as women, that comes to about $10.4\%$. This figure is relatively better than the figures for participation of women in most surveys in the years $2000-2003$. In the survey few questions relating to women's issues have been asked and so the data cannot be used to answer pressing questions on the participation of women in FOSS. For an idea of the issues, the reader is referred to \cite{AIS2011}.

The percentage of women with an university education is $81.4\%$, while the corresponding figure for the whole dataset is $72\%$. Is this suggestive of higher barriers of entry? Fig.1 says more about this aspect.

\begin{figure}[hbt]
\centering
\includegraphics[width=6.0cm]{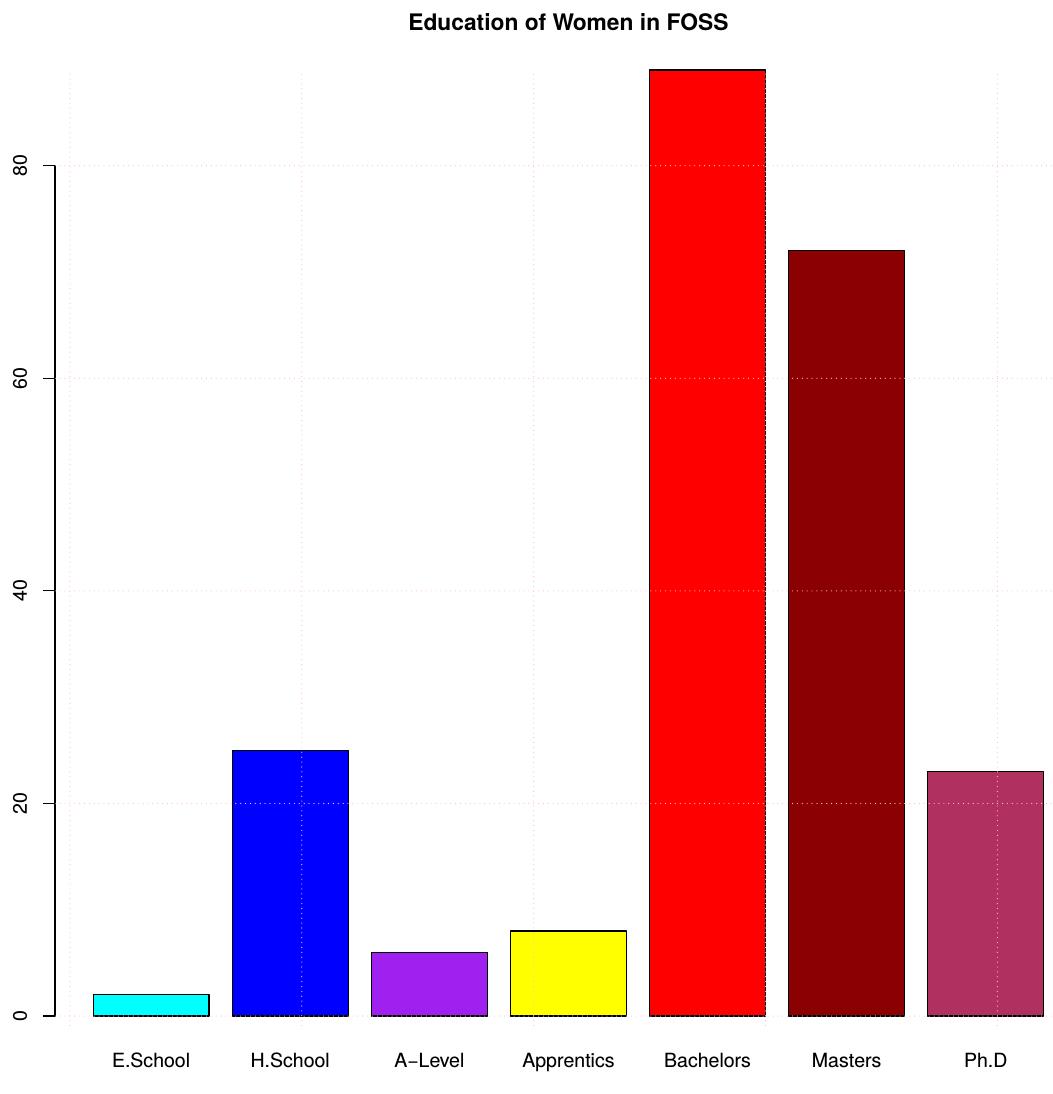}
\caption{Education Level of Women}
\end{figure}

$76\%$ of women contributors contribute code, while the figure for the whole dataset is $72\%$. This is a good development as women have been known to stick to traditional roles like documentation in the FOSS world. Most of them ($81\%$) believe that FOSS development is not self exploitive in any sense. This is reflected in Fig.2

\begin{figure}[hbt]
\centering
\includegraphics[width=6.0cm]{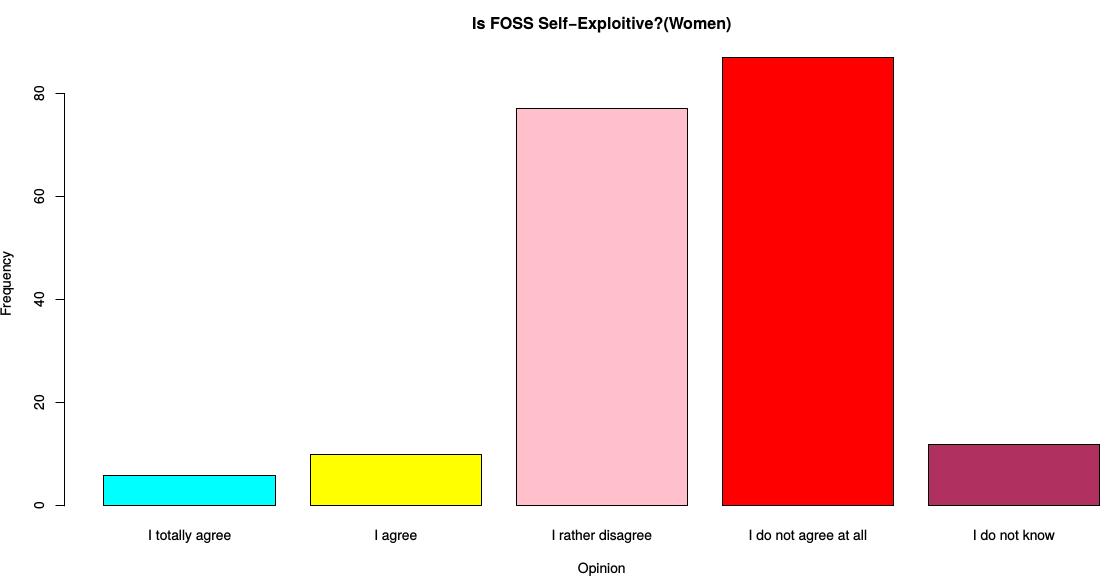}
\caption{Contribution=Self Exploitation?}
\end{figure}

Believing in FOSS being non self exploitive may be expected to rhyme with belief in community development models and collaborating with people with similar interests. The dataset actually says so.

The following plot (Fig.3) suggests that most woman tend to start contributing to free software in their twenties.

\begin{figure}[hbt]
\centering
\includegraphics[width=6.0cm]{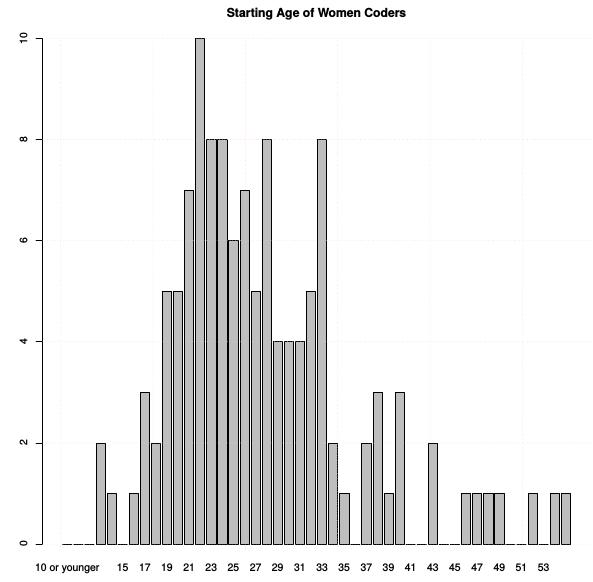}
\caption{Age at Time of First Contribution}
\end{figure}

\subsubsection*{MCA}

Multiple correspondence analysis \cite{FSJ2011} over the three subsets corresponding to outlook, programming and development environment for women are investigated next. The library \texttt{FactoMineR} was used for all this.

Using MCA methods on women participant's basic outlook of the free software and OSS world, the associated eigen values had the following form and alluded to reasonably strong influence of the first two dimensions:

\begin{verbatim}
#Outlook of Women
> head(mcafemphilfse$eig)
  eigenvalue  %var     cum% var
dim 1 0.19    6.8        6.8
dim 2 0.08    3.0        9.9
dim 3 0.06    2.2       12.0
dim 4 0.05    2.0       14.1
dim 5 0.05    2.0       16.0
dim 6 0.05    1.9       17.9
\end{verbatim}

The HCPC procedures indicate about $15$ very important questions for characterization of outlook in decreasing order of importance (though the first five can suffice for many indicators). These questions, in order, are about
\textsf{involvement in proprietary s/w development, age, year of getting involved, question of directly earning from FOSS, whether the need to learn new skills was a motivator for getting into free software world and a valid reason for continuing to do so}. Part of the relevant sequence is Q0024, Q0022, Q0025, Q0021, Q0035, Q0023, Q0030\_2, Q0034, Q0032\_2, Q0029, Q0028, Q0031\_2, Q0031\_9, Q0033\_3, Q0032\_3, Q0030\_5. This suggests that negative questions can be good predictors of overall outlook. 

\begin{verbatim}
#Outlook of Women
> hcpcfemphilfse$desc.var
$test.chi2
              p.value  df
Q0024    1.345859e-78  10
Q0022    1.032740e-53 185
Q0027    2.432245e-48  85
Q0025    6.112861e-46  10
Q0021    3.012689e-41 100
Q0035    6.556618e-39  40 
\end{verbatim}

Using MCA techniques on women contributor's views on development environment and challenges lead to high eigenvalues for the first three dimensions

\begin{verbatim}
#Views of Women on Development
> head(mcafemvewfse$eig)
eigenvalue  var%   cum% var
dim 1 0.21  9.2    9.2
dim 2 0.14  6.5    15.7
dim 3 0.08  3.6    19.3
dim 4 0.06  2.6    21.9
dim 5 0.05  2.3    24.2
dim 6 0.05  2.2    26.4
\end{verbatim}

The number of questions required to obtain the position of women contributors on development issues include at least twenty questions and their answers. This conclusion is based on application of the \texttt{hcpc()} function on the MCA object. The questions in descending order of importance are: Q0057\_1, Q0057\_4, Q0057\_2, Q0057\_6, Q0058\_8, Q0057\_5,  Q0058\_11, Q0057\_7, Q0058\_9, Q0057\_3, Q0058\_12 and Q0058\_10. Which in words relate to challenges relating to \emph{importance of software stability by reduction of bugs, making software suitable for mobiles, etc, improving feature set, attracting users, leveraging the power of the leader/main company, reduction of time between releases, communication inside projects, getting users involved in contribution, getting enough funds, improving documentation, upstream communication and legal issues}.  An extract is shown below:

\begin{verbatim}
> hcpcfemvewfse$desc.var
$test.chi2
              p.value df
Q0057_1  3.158089e-61 21
Q0057_4  2.622154e-59 18
Q0057_2  6.476614e-59 21
\end{verbatim}

Associated clusters involving both response of users and categories are part of the above command. In the specific case the first two clusters are less relevant as they referred NA responses. The other two (in the selection) are way better. But all four help in categorizing users in terms like \emph{the ones who responded with A1 to question X1 and so forth belong to a group who are likely to answer B to question Z}. But do the categories matter? In the first author's opinion, they do. But as the details are cumbersome, these will be published separately. Fig.4 shows the 3-D clustering associated with HCPC method and the effectiveness of dimensions is reflected in Fig. 5.

\begin{figure}[hbt]
\centering
\includegraphics[width=6.9cm]{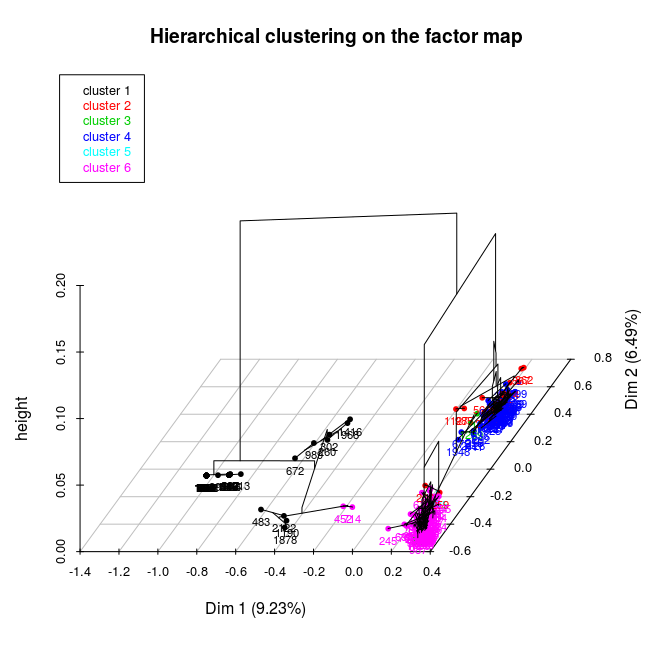}
\caption{Views on Development Environment}
\end{figure}

\begin{figure}[hbt]
\centering
\includegraphics[width=6.9cm]{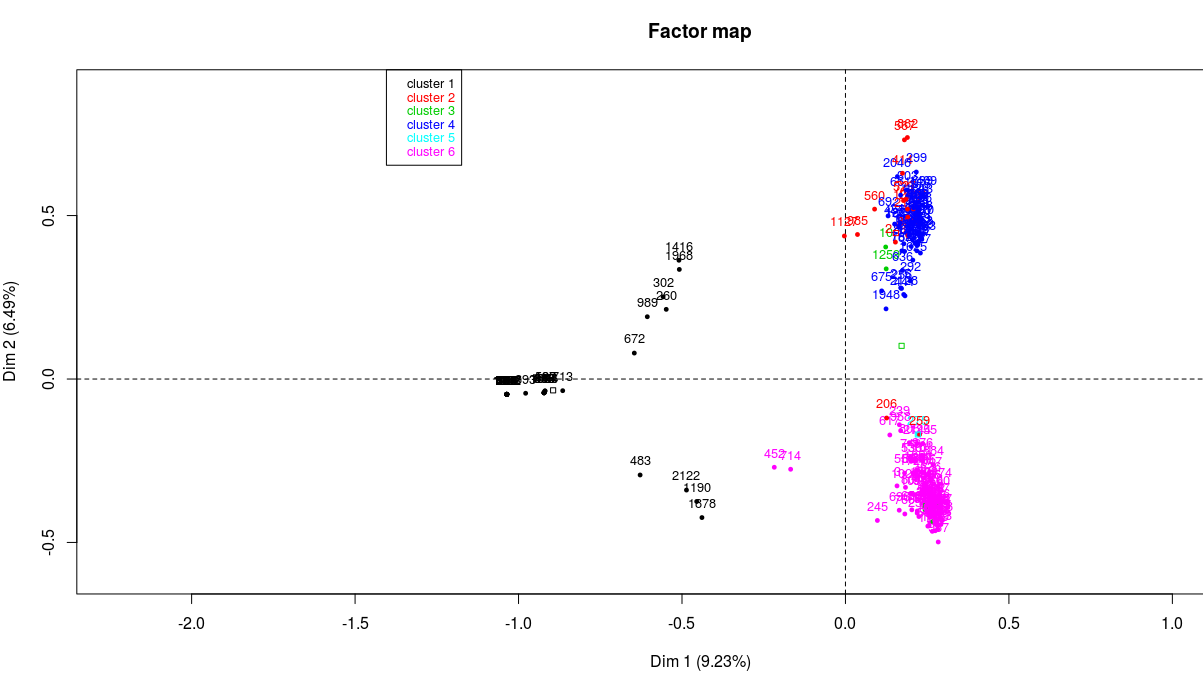}
\caption{Views on Development Environment}
\end{figure}

From women contributor's response to programming related question, good dimensions are extractable by the MCA approach. The eigen values below say as much.

\begin{verbatim}
#Women on Programming
> head(mcafemprogfse$eig)
     eigenvalue  %var    cum% var
dim 1 0.18       12.73     12.73
dim 2 0.09        6.52     19.25
dim 3 0.06        4.24     23.49
\end{verbatim}

Unlike for views on issues, responses to questions on programming languages are expected in binary form. Using hcpc, a large number of questions (nearly 30) turn out to be important and orders are assignable on them. The important programming languages turn out to be Modula, Ada, Smalltalk, Tcl, Pascal, HTML and so on with C in the tenth position. But this is due to the nature of the questions in the subset. Clustering works as per expectations with top programming languages (as per the recent IEEE classification) forming a cluster.

\begin{flushleft}
\textbf{Remarks} 
\end{flushleft}

In this research, the FOSS'2013 survey is contextualized and critically examined in the light of other available data and surveys on women in free software, a number of relatively simpler statistics have been computed, stress has been laid on women in free software and their views on free software development environments and their outlook has been elucidated using MCA and HCPC methods. Smaller partitions of the dataset using lesser number of features are also of interest and would be examined in a supportive study. This study would also be useful for improving related survey methodologies and design.

\bibliographystyle{IEEEtran}
\bibliography{biblioam06092016.bib}
\end{document}